\documentclass[10pt, conference, a4paper]{IEEEtran}
\IEEEoverridecommandlockouts

\usepackage{times}
\usepackage{epsfig}
\usepackage{graphicx}
\usepackage{amsmath}
\usepackage{amssymb}
\usepackage[super]{nth}
\usepackage[warn]{textcomp}
\usepackage{mathtools}
\usepackage[T1]{fontenc}
\usepackage[pagebackref=true,breaklinks=true,letterpaper=true,colorlinks,bookmarks=false]{hyperref}

\def\BibTeX{{\rm B\kern-.05em{\sc i\kern-.025em b}\kern-.08em
    T\kern-.1667em\lower.7ex\hbox{E}\kern-.125emX}}
\begin{document}

\title{Multi-scale Feature Aggregation for Crowd Counting}

\author{Xiaoheng Jiang, Xinyi Wu, Hisham Cholakkal, Rao Muhammad Anwer, Jiale Cao,\\
Mingliang Xu$^*$, Bing Zhou, Yanwei Pang, Fahad Shahbaz Khan}

\maketitle


\begin{abstract}
Convolutional Neural Network (CNN) based crowd counting methods have achieved promising results in the past few years. However, the scale variation problem is still a huge challenge for accurate count estimation. 
In this paper, we propose a multi-scale feature aggregation network (MSFANet) that can alleviate this problem to some extent.
Specifically, our approach consists of two feature aggregation modules: the short aggregation (ShortAgg) and the skip aggregation (SkipAgg). The ShortAgg module aggregates the features of the adjacent convolution blocks. Its purpose is to make features with different receptive fields fused gradually from the bottom to the top of the network. The SkipAgg module directly propagates features with small receptive fields to features with much larger receptive fields. Its purpose is to promote the fusion of features with small and large receptive fields. Especially, the SkipAgg module introduces the local self-attention features from the Swin Transformer blocks to incorporate rich spatial information.
Furthermore, we present a local-and-global based counting loss by considering the non-uniform crowd distribution. Extensive experiments on four challenging datasets (ShanghaiTech dataset, UCF\_CC\_50 dataset, UCF-QNRF Dataset, WorldExpo'10 dataset) demonstrate the proposed easy-to-implement MSFANet can achieve promising results when compared with the previous state-of-the-art approaches.

\end{abstract}

\section{Introduction}
\label{intro}


The past decade has witnessed great progress in crowd counting. The counting strategies have devolved gradually from detection-based methods~\cite{li2008estimating,zeng2010robust,lin2010shape,shami2018people} to density regression-based methods~\cite{lempitsky2010learning,zhang2016single,li2018csrnet} since the wild crowd scenes often bring challenges like heavy congestion and large scale variation. Nowadays, convolutional neural network (CNN) based counting methods~\cite{KONG2020105927,LI2020106485,GUO2021106691,onoro2016towards,sam2017switching,liu2019adcrowdnet,liu2020crowd} have attracted a lot of attention due to their ability to extract rich hierarchical features. CNN based density estimation methods project the features of an input image onto a density map, on which the total people count can be determined by summing up the densities. To handle the scale variation problem, these methods generally adopts several techniques such as exploring multi-scale feature~\cite{zhang2016single,KONG2020105927,guo2019dadnet,sindagi2017generating}, embedding dilation convolution~\cite{li2018csrnet}, utilizing attention mechanism~\cite{hossain2019crowd,zhang2019relational,zhang2019attentional,liu2019recurrent}, and so on. However, the counting performance is still far from satisfactory.

\begin{figure}[hbt]
	\centering
		\includegraphics[scale=0.35]{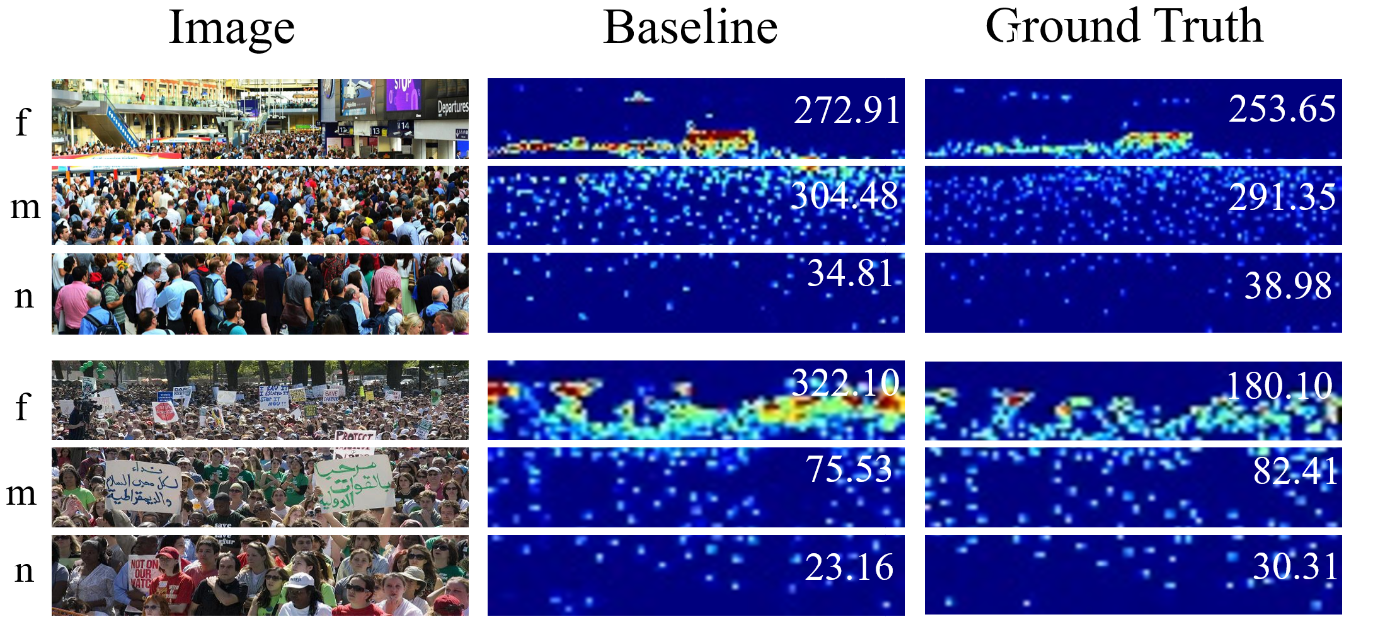}
	\caption{The crowd scenes can be coarsely divided into three regions according to the distances from the camera: far (f), middle (m) and near (m). The counting network usually performs differently in the three regions, with larger count estimation errors in the regions far away from the camera. The f and m regions are typically characterized by heavy people occlusion and high densities.}
	\label{fig1}
\end{figure}

The data-driven CNN counting network usually suffers from non-uniform performance in different regions, especially resulting in relatively larger estimation errors in high-density regions where people are small and are heavily occluded. 
Figure.~\ref{fig1} shows that the count estimation errors of the given input crowd scenes mainly come from the high-density regions far away from the camera. It is known that the deep convolutional neural network has a hierarchy of features. The features close to the input of the network contain low-level features with local receptive fields, such as edge information, texture information, etc. The features close to the output of the network contain high-level semantic features with large or global receptive fields. The high-level features contain context information that is important to scene content analysis. 
However, in the crowd counting task, the heads of the high-density regions far from the camera are usually very small and the corresponding local low-level features are also important to the density estimation. When increasing the network depth, the detailed information contained in the local low-level features gradually decreases.  Hence, the traditional CNN based counting network does not work well in regions with congested small heads. 

In this paper, we aim to ensure the features projected to the density map contain rich multi-scale information about the crowd scenes, especially the high-density regions. To this end, we present a multi-scale feature aggregation convolutional network named MSFANet. 
The core idea of MSFANet is to enhance the feature representation ability by aggregating low-level features with local receptive fields and high-level semantic features with large receptive fields.
MSFANet densely aggregates the features of various scales at multiple stages by utilizing two feature aggregation modules.
The first module is the short aggregation module named as ShortAgg. ShortAgg aggregates the features of adjacent scales and sends them into the next convolution block. A dense insertion of the ShortAgg modules helps the local low-level features flow gradually from the bottom to the top of the network.
The second module is the skip aggregation module named SkipAgg. SkipAgg directly combines the local low-level features with the high-level features that have much larger receptive fields. Inspired by the recent great success that the vision transformers have made in tasks such as object classification~\cite{2020CarionEnd} and object detection~\cite{2020An}, we introduce the Swin-Transform blocks~\cite{2021Swin} to extract low-level global features and propagate them to high-level CNN features in the SkipAgg module. 

Furthermore, we present a loss function based on the combination of the global and local density estimation losses. The global loss part is the standard Euclidean loss between the whole estimated density map and the corresponding ground truth of an input image. The local loss part is called pooling loss (PLoss) which utilizes a locality-aware loss (LA-Loss) kernel to aggregate all the local losses in the pooling manner. The PLoss helps attenuate the loss differences between regions of different crowd densities and benefits the generalization of the learned counting network.

The major contributions of this work can be summarized as follows:
\begin{enumerate}
\itemsep=0pt
\item We introduce a multi-scale feature aggregation convolutional network (MSFANet) that utilizes two feature aggregation paths to propagate local low-level features from the bottom to the top of the network both gradually and directly, which helps strengthen the multi-scale representation ability of the output features.
\item We present a short aggregation module (ShortAgg) that merges features of the adjacent scales and a skip aggregation module (SkipAgg) that merges features with large scale differences.
\item We present a combined density estimation loss that considers both the global and local density estimation losses, which alleviates the bias caused by the non-uniform crowd distributions of low and high densities.
\item Extensive experimental results on four challenging crowd datasets demonstrate that our proposed approach performs favorably against state-of-the-art crowd counting methods.

\end{enumerate}

\section{Related Works}
\begin{figure*}[tbh]
	\centering
	\includegraphics[scale=0.36]{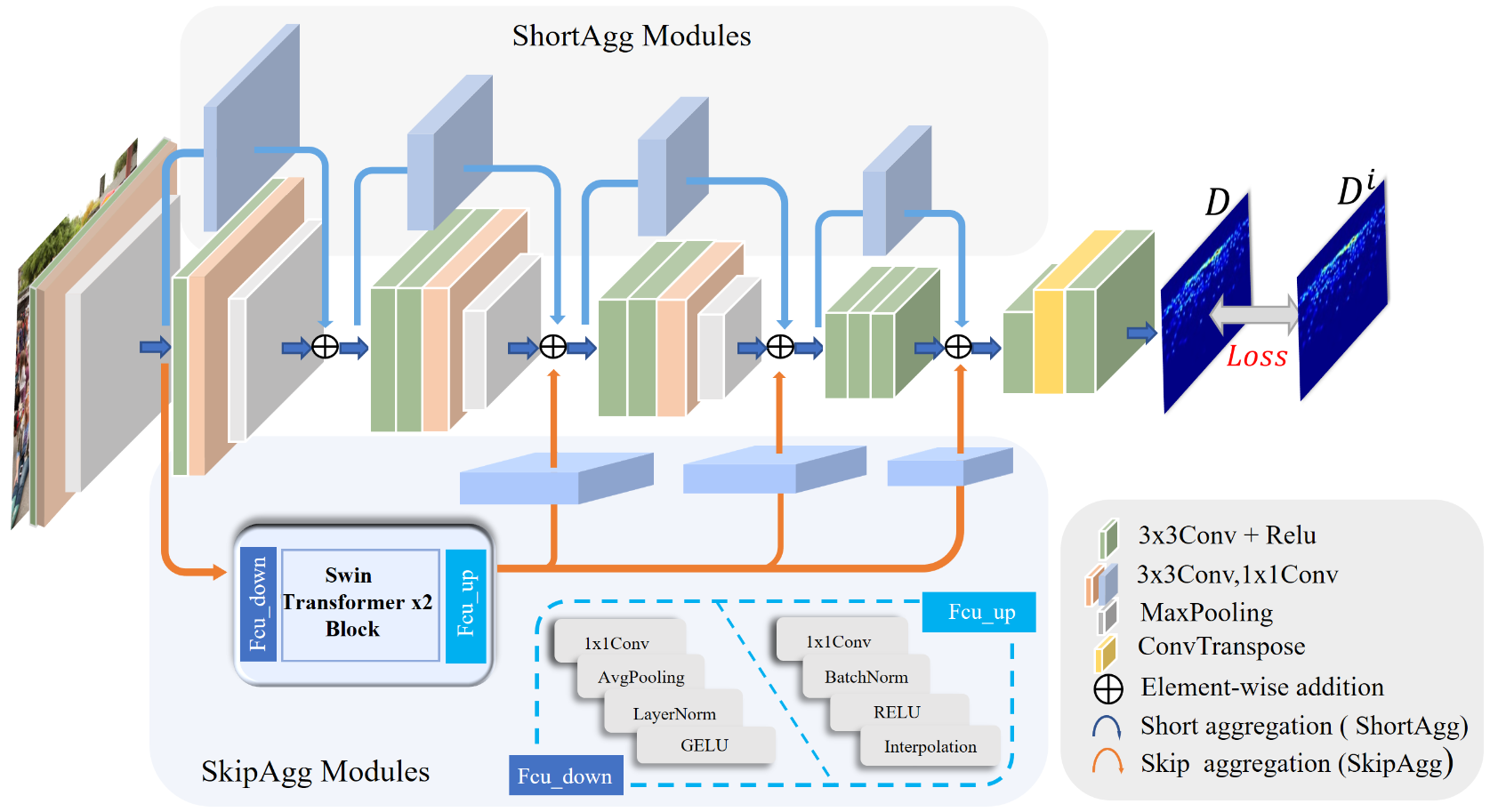}
	\caption{The architecture of the proposed multi-scale feature aggregation network (MSFANet). It consists of the VGG-16 backbone, the short feature aggregation (ShortAgg) path, and the skip feature aggregation (SkipAgg) path. ShortAgg fuses features of the adjacent scales and SkipAgg directly propagates the local low-level transformer features from the bottom to the top of the network. ShortAgg and SkipAgg together make the output features rich in multi-scale information, which helps promote the performance of the counting network.}
	\label{fig2}
\end{figure*}

\subsection{Multi-scale Feature Fusion}
These kinds of approaches utilize multi-scale features of the input image to address the problem of object scale variation.
Early on, Zhang  \textit{et al.}~\cite{zhang2016single} proposed the multi-column convolutional neural network (MCNN) to capture multi-scale features by utilizing convolution kernels of different sizes in three branches.
Onoro  \textit{et al.}~\cite{onoro2016towards} develop the Hydra CNN that captures multi-scale information by using image patches of one image pyramid. As a result, the Hydra network can robustly estimate crowd density in various crowded scenarios.
The Switch-CNN, proposed by Sam \textit{et al.}~\cite{sam2017switching}, consists of three CNN regressors that have the same structure as MCNN. However, it uses a switch classifier to select the best regressor for different image patches.
Sindagi  \textit{et al.}~\cite{sindagi2017generating} presented a contextual pyramid CNN (CP-CNN) that incorporates the global and local context to produce high-quality crowd density maps.
The TDF-CNN, proposed by Sam  \textit{et al.}~\cite{sam2018top}, delivers the top-down feedback information as a correction signal to the bottom-up network to correct the density prediction. The bottom-up CNN has two columns with different receptive fields to regress the density maps.
%
%
The DADNet, proposed by Guo \textit{et al.}~\cite{guo2019dadnet}, generates high-quality density maps by using multi-scale dilated attention to learn context cues of multi-scale features.

\subsection{Attention Mechanism}
These kinds of methods usually utilize the  attention mechanism to highlight the regions containing the crowd and filter out the noises of the background.
Liu \textit{et al.}~\cite{liu2019recurrent} utilized attention models to  detects ambiguous image region recurrently and zooms it into high resolution for re-inspection.
Hossain \textit{et al.}~\cite{hossain2019crowd} proposed a scale-aware attention network that uses the global and local attention mechanism to automatically select features of the appropriate scale.
Zhang \textit{et al.}~\cite{zhang2019relational} proposed a relational attention network that is composed of an attention module and a relation module. The attention module utilizes a local self-attention (LSA) and global self-attention (GLA) mechanism to capture long and short-range interdependence information. The relation module fuses the information obtained by LSA and GSA to obtain robust features.
The attentional neural field network(ANF), proposed by Zhang \textit{et al.}~\cite{zhang2019attentional}, utilizes a non-local attention mechanism to expand the receptive field, captures long-range dependencies, and enhances the representation of multi-scale features in the network.

\subsection{Vision Transformer}
Inspired by the success that transformers have made in the field of natural language processing, researchers begin to explore the self-attention mechanism for visual tasks such as image classification~\cite{2020An,2021Learning,2021Swin}, object detection~\cite{2020CarionEnd,2020Deformable,2020zhangEnd,2020SunRethinking}, object segmentation~\cite{2020WangEnd,2020MaX}, etc.
Specifically, ViT~\cite{2020An} directly applies the transformer encoder to accomplish the classification task. It interprets an image as a series of patches and takes the linear embedding sequence of these image patches as input of the transformer network. 
DETR~\cite{2020CarionEnd} is an end-to-end CNN and transformer-based object detection network. It first utilizes a CNN backbone to extract features and then utilizes a stack of transformer blocks to generate features for object detection.
Based on DETR, Deformable DETR~\cite{2020Deformable} uses deformable convolution to sparsely sample the features and focuses only on the key positions, which reduces the computational complexity and improves the convergence speed.
Similarly, SETR~\cite{2020ZhangRethinking} proposes to replace the CNN encoder with a pure transformer structure encoder, which can fully explore the segmentation capabilities of ViT~\cite{2020An}.
Wang \textit{et al.}  proposed a video instance segmentation framework named VisTR~\cite{2020WangEnd}, which directly treats the video instance segmentation task as a direct end-to-end parallel sequence decoding and prediction problem.
Liu \textit{et al.} proposed a hierarchical transformer, named Swin Transformer~\cite{2021Swin}, which improves the efficiency by restricting the self-attention computation within the non-overlapping local windows and allowing the cross-window connection.

In this paper, we utilize Swin Transformer blocks to capture low-level feature information and propagate them to top CNN layers to accomplish the fusion of transformer features and CNN features, which help promote the multi-scale feature representation ability of the counting network.

\section{Our Approach}
The overall architecture of the proposed MSFANet method is illustrated in Figure~\ref{fig2}. The VGG-16 model is chosen as the backbone due to its simple and powerful architecture and its success in many vision tasks. And two additional feature propagation paths go from the bottom of the backbone to the top of it via two feature aggregation modules. They are called the short aggregation module (ShortAgg) and the skip aggregation module (SkipAgg), respectively.
In this section, we first introduce the ShortAgg module and SkipAgg module and then present the pooling loss (PLoss) 

\subsection{ShortAgg module}
\begin{figure}[tbh]
	\centering
	\includegraphics[scale=0.38]{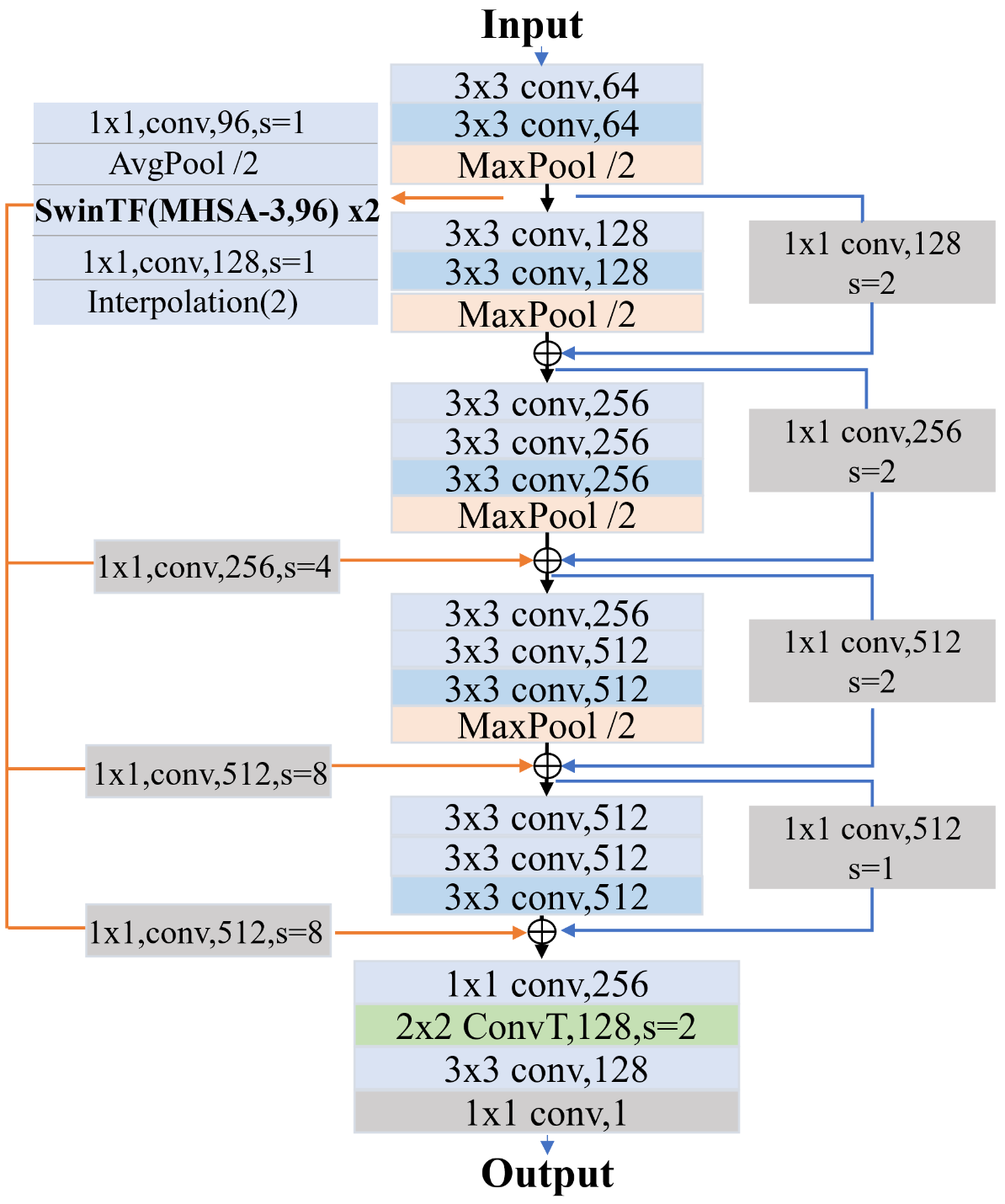}
	\caption{The configuration of the proposed MSFANet method. The middle part is the VGG-16 backbone, the left part is the SkipAgg, and the right part is the ShortAgg. The density regressor is at the end of the fused output features.}
	\label{fig3}
\end{figure}
The purpose of ShortAgg is to aggregate the features that have an adjacent receptive field (i.e., features of adjacent scales ). ShortAgg combines the features of the current convolution block and the preceding one and takes them as input of the subsequent convolution block. The structure of ShortAgg is similar to the residual connection proposed by He \textit{et al.}~\cite{he2016deep}. Residual connection aims to deal with the problem of vanishing gradients whereas ShortAgg is to promote feature fusion. The ShortAgg module is defined as: 
\begin{equation}Y_{SH}^{i}\;=\;P\{F(X_i,w)\}+ w_s\ast X_i, \end{equation}
$ X_i $ and $Y^{i}$ represent the input and the output features of the $i-th$ convolution block, respectively. The function $F$ represents the feature mapping which consists of two or three convolutional layers in a block, and $P$ denotes the maximum pooling operation. For example, $ F=\omega2\ast ((\sigma(\omega1\ast X+b1))+b2) $  indicates a two-layer convolution block, where $\sigma$ denotes the activation function ReLU, $\omega1$, $ b1$ and $\omega2$ , $ b2$ represent the weights and biases of each corresponding convolutional layer, respectively.
The $'+'$ operation indicates the ShortAgg module that accomplishes the feature aggregation between $X$ and $F$. A simple convolution is operated on $X$ to ensure that the dimensions of $X$ and $F$ are the same.

The configuration of the ShortAgg is presented in Figure~\ref{fig3}. The right part with blue connection lines represents the established ShortAgg, and the gray rectangles are the $1\times1$ convolutional layers introduced by ShortAgg, $s$ represents the stride of the convolution. We utilize the convolution with a kernel of $1\times1$ and $s=2$ in the ShortAgg to ensure that the dimensions of $X$ and $F$ are the same. 
We divide the first thirteen layers of the VGG-16 model into five blocks and each block performs a maximum pooling operation at the end except for the last block. There are two $3\times3$ convolution layers in the first two blocks, three $3\times3$ convolution layers in the last three blocks. Starting from the second block, each block adopts a ShortAgg module.

\subsection{SkipAgg module}
Local low-level CNN features contain detailed information of objects and play an important role in detecting small objects. In the crowd counting task, the human heads in the far-away regions are usually very small and the corresponding features are easy to lose at the top of the network.
Therefore, we further propose the skip aggregation module (SkipAgg) to directly propagate the local low-level features to the top layers of CNN. Specifically, SkipAgg is to aggregate the input and the output of multiple stacked convolutional blocks to achieve cross-scale feature fusion.

As shown in Figure~\ref{fig3}, we introduce the Swin-Transformer block to extract local low-level features and directly propagate them to high-level features to obtain fused Transformer-CNN features. Swin Transformer is adopted due to that it can effectively model the local information via the multi-head self-attention module within the patch windows.
And then the $1\times1$ convolutions with different strides are used to make the resolutions of the transformer features adapted to the high-level CNN features. The SkipAgg module is defined as:
\begin{equation}
	Y_{SK}^{i+n-1}=n * P\left\{F(X_i,w)\right\}+w_s\ast X_i,
\end{equation}
where $n$ represents the number of blocks that need to be skipped, $X_i $ represents the input of the $i-th$ convolution block, and $Y_{SK}^{i+n-1}$ represent the output of the $(i+n-1)-th$ convolution block.
In our MSFANet, the values of n are set to 2, 3, and 4, respectively. When $n = 2$, the output of the first block connects to the output of the third block (i.e., skip two blocks). Similarly, $ n=3$ denotes that the output of the first block connects to the output of the fourth block (i.e., skip three blocks).
As a result, the low-level transformer features generated in the first block can be directly propagated to the topper CNN layers and be fused with the high-level CNN features that have large receptive fields. This can further make the output features of the network contain more information about small objects.

\begin{figure}[tbh]
	\centering
	\includegraphics[scale=0.34]{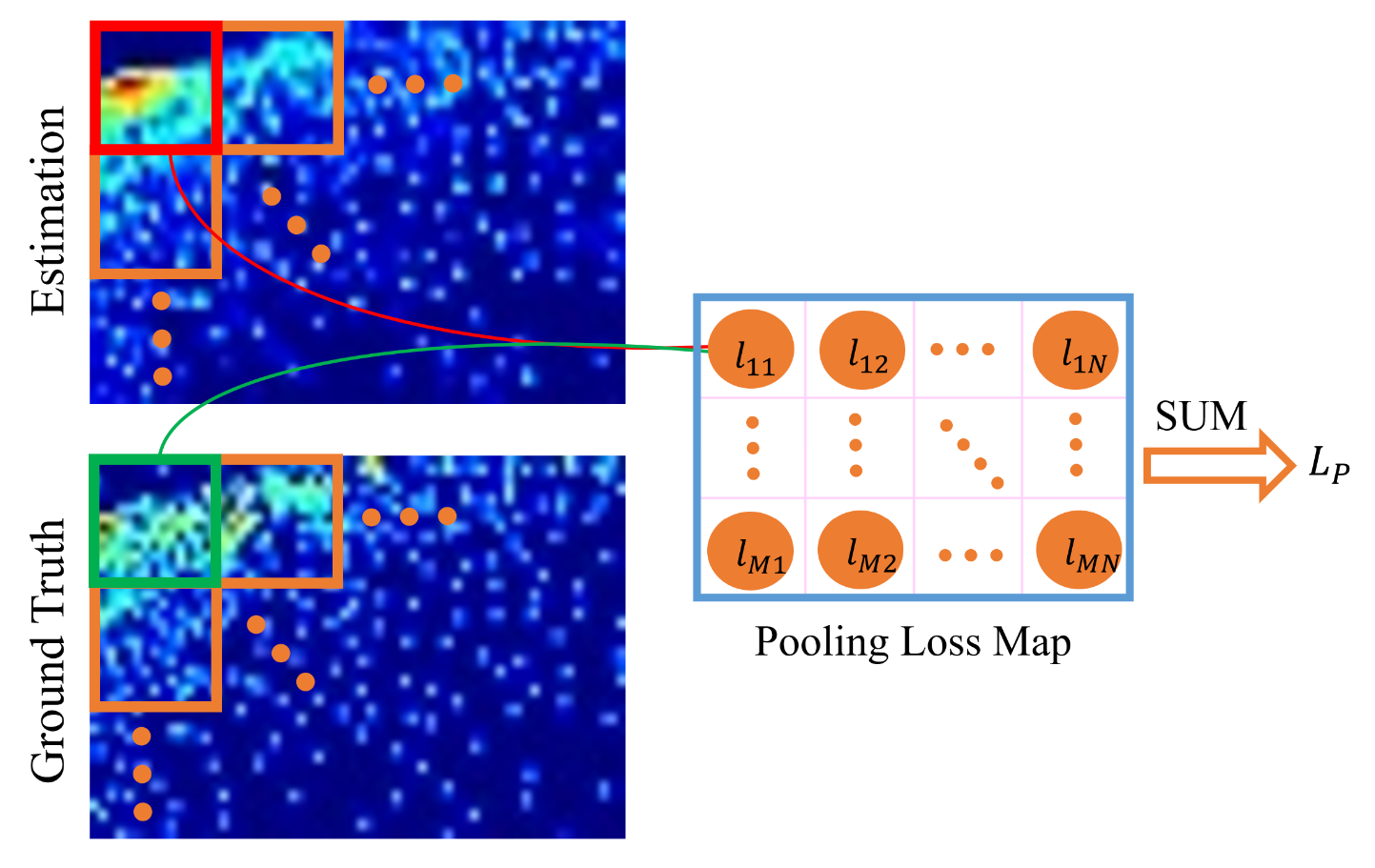}
	\caption{The demonstration of the proposed pooling loss. The  locality-aware losses are gathered in the pooling manner to obtain the whole density estimation loss.}
	\label{fig4}
\end{figure}

The configuration of SkipAgg is presented in Figure~\ref{fig3}. The left part shows the SkipAgg path. The transformer module contains a stack of layers, including one $1\times1$ convolutional layer with 96 channels, one average pooling layer with a stride of 2, two Swin Transformer blocks, one convolutional layer with 128 channels, and one interpolation layer. 
And then three  $1\times1$ convolutional layers with strides of 4, 8, 8 are utilized to make the transformer features adapted to three different CNN blocks, respectively. The element-wise addition operation is adopted to merge the corresponding features.

It is noted each the SkipAgg module also fuses with the ShortAgg module, which strengthens the multi-scale feature aggregation and promotes feature propagation from the bottom of the network to the top layers. The whole process can be defined as:
\begin{equation}
	Y^{i+n-1}=	Y_{SH}^{i} + Y_{SK}^{i,i+n-1},
\end{equation}
where the output features $Y^{i+n-1}$ of the $(i+n-1)-th$ convolution block contain features from the main backbone, the ShortAgg path, and the SkipAgg path.

\subsection{Loss Function}
Most of the previous CNN-based crowd counting works adopt standard Euclidean distance between the whole estimated density map and the whole ground truth density map of input as the loss during back propagation to guide the next training direction of the model. The traditional Euclidean distance loss is then defined as follows:
\begin{equation}
	L_E = \frac{1}{K}\sum_{i=1}^{K}||D(X^i;\Theta) -D^i||_{2}^{2},
\end{equation}

where $ K $ is the number of samples used for training. $D(X^i;\Theta) $ and $ D^i $ represent the estimated density map and the ground truth density map for an input sample $ X^i $ respectively.$ \Theta $ is trainable parameters in the counting network.This kind of loss averages the errors caused by different local density distributions. In other words, it ignores the individual contribution of each local density distribution.

To overcome this issue, we present a pooling loss (PLoss) that uses a new loss kernel named locality-aware loss (LA-Loss) to calculate each loss value in the pooling manner. The pooling loss is presented in Figure~\ref{fig4}.  The LA-Loss is defined as follows:
\begin{equation}
	l_{mn}^{i}( \Theta) = \frac{||D_{mn}(X^i;\Theta) -D_{mn}^i||_{2}^{2}}{SUM(D_{mn}^i)+1},
\end{equation}
where \textit{m} and \textit{n} are the indices of the pooling window in the vertical and horizontal directions, respectively. And $ D_{mn}(X^i;\Theta) $ and $ D_{mn}^i $ are the corresponding windows on the predicted density map and the ground truth density map for sample $ X^i $, respectively. $ SUM(D_{mn}^i) $ sums the densities of each pooling window $ D_{mn}^i $ to get the number of people in it. $ SUM(D_{mn}^i)+1$ prevents the division by zero.

And then the total loss $ l^{i}( \Theta) $ of the whole density map for the sample $ X^i $ is obtained by gathering all the individual losses of the map.
Finally, the proposed PLoss is used for back-propagation during training and is computed as follows:
\begin{equation}
	L_P = \frac{1}{K}\sum_{i=1}^{K}l^{i}( \Theta).
\end{equation}
And the final density estimation loss function is defined as follows:
\begin{equation}
	L_T = \alpha L_E + L_P.
\end{equation}
where $\alpha $ is the weighting coefficient to balance the standard Euclidean distance loss and PLoss. In our experiments, we set $ \alpha $ to 0.1.

\section{Experiments}
In this section, we demonstrate the effectiveness of the proposed method on four challenging crowd datasets, including the ShangehaiTech Part A Dataset~\cite{zhang2016single}, UCF\_CC\_50 dataset~\cite{idrees2013multi}, UCF-QNRF dataset~\cite{idrees2018composition}, and WorldExpo'10 dataset~\cite{zhang2015cross}. We first introduce the four datasets and then present implementation details and the evaluation metrics used in the experiments. Finally, we present the experimental results and analysis. The implementation of our method is based on the PyTorch framework.

\subsection{Datasets}
\textbf{ShanghaiTech Part A Dataset.} This dataset is introduced by Zhang \textit{et al.}~\cite{zhang2016single} and contains 482 images collected from the website randomly. It is divided into the training and test subsets, with 300 images for training and 182 images for the test, respectively. In addition,The dataset has many complex scenes with the challenge of population-scale change.

\textbf{UCF$\_ $CCF$ \_ $50 Dataset.} This dataset introduced by Idrees \textit{et al.}~\cite{idrees2013multi} presents many challenges. It contains only 50 images randomly collected from the website. The scenes are very crowded with 1280 individuals on average in each image and 63974 annotations in total. And it contains a wide range of densities with the head number per image changing largely from 94 to 4543. 

\textbf{UCF-QNRF Dataset.} This is a large dataset introduced for crowd counting~\cite{idrees2018composition}. It is collected from the web and has a large variation in perspective, image resolution, and crowd density. It contains 1535 images with a total of 1,251,642 annotations.It is divided into the training and test subsets, with 1,201 images for training and 334 images for the test, respectively. 

\textbf{WorldExpo'10 dataset.} This dataset is collected and published by Zhang \textit{et al.}~\cite{zhang2015cross} to solve the problem of rowd counting in cross-domain scenes.It comes from the video sequences captured by 108 surveillance cameras of the 2010 Shanghai World Expo.There are 1132 annotated video sequences with 199,923 annotated pedestrians at the centers of their heads. The dataset is divided into a training set with 103 scenes and a test set with 5 scenes.

\subsection{Implementation details}
\textbf{Ground truth generation.}
We generate the ground truth density maps by using a Gaussian kernel to blur the head annotations. The Gaussian kernel is normalized to sum to one so that summing the density map gives the crowd count. In our experiments, we use a fixed spread Gaussian to generate the density maps. Since the output density map is $ 1/8 $ size of the input image, we resize the ground truth density map of the original input image as a loss calculation needs to one-eighth of the input, and the processing operation is down sampling. 

\textbf{Data Augmentation.}
The training data is augmented by randomly cropping image patches of $224\times 224$ pixels in one image. On UCF-QNRF Dataset, the patch size is set to $512\times 512$. We also scale the original image with scales of 0.75 and 1.25 and mirror each image patch horizontally.

\textbf{Training.}
We initialize the front part (the first 13 convolutional layers ) of MSFANet by adopting a well-trained VGG-16 model. For the rest layers of the network, we initialize the weight values using a Gaussian of 0.01 standard deviation. The whole network is trained in an end-to-end manner. We train the MSFANet with $L_T $ by adopting the Adma optimization algorithm.

\subsection{Evaluation Metrics}
\label{secMetric}
To uniformly evaluate the performance of the algorithm, we use the general algorithm evaluation standard Mean Absolute Error (MAE) and Mean Square Error (MSE) in the crowd counting task to evaluate our method.MAE and MSE are defined as follows:
\begin{equation}
	MAE = \frac{1}{K}\sum_{i=1}^{K}|C_i-\hat{C_i}|,
\end{equation}
\begin{equation}
	MSE =\sqrt{\frac{1}{K}\sum_{i=1}^{K}||C_i-\hat{C_i}||^{2}},
\end{equation}
where $ K $ is the number of images in the test set, $ C_i $ is the ground truth crowd count of the \textit{i}-th image, and $ \hat{C_i} $ is the predicted crowd count which integrates the corresponding generated density map. 

\subsection{Ablation study}
\label{secAblation}
We conduct an ablation study on ShanghaiTech Part A dataset~\cite{zhang2016single} to evaluate the effectiveness of the proposed ShortAgg module, SkipAgg module, and PLoss.  The experimental results are shown in Table~\ref{tableAblation}.

\textbf{Baseline.} The baseline model contains a VGG-16 model by removing the fifth max-pool layer and the fully connected layers and adds three additional convolution layers and one transposed convolution layer to regress the final density map. The model is trained in an end-to-end way with the standard Euclidean distance loss. The MAE is 65.22 and the MSE is 102.18 on the ShanghaiTech Part A dataset~\cite{zhang2016single}.

\begin{figure}[tbh]
	\centering
	\includegraphics[scale=0.26]{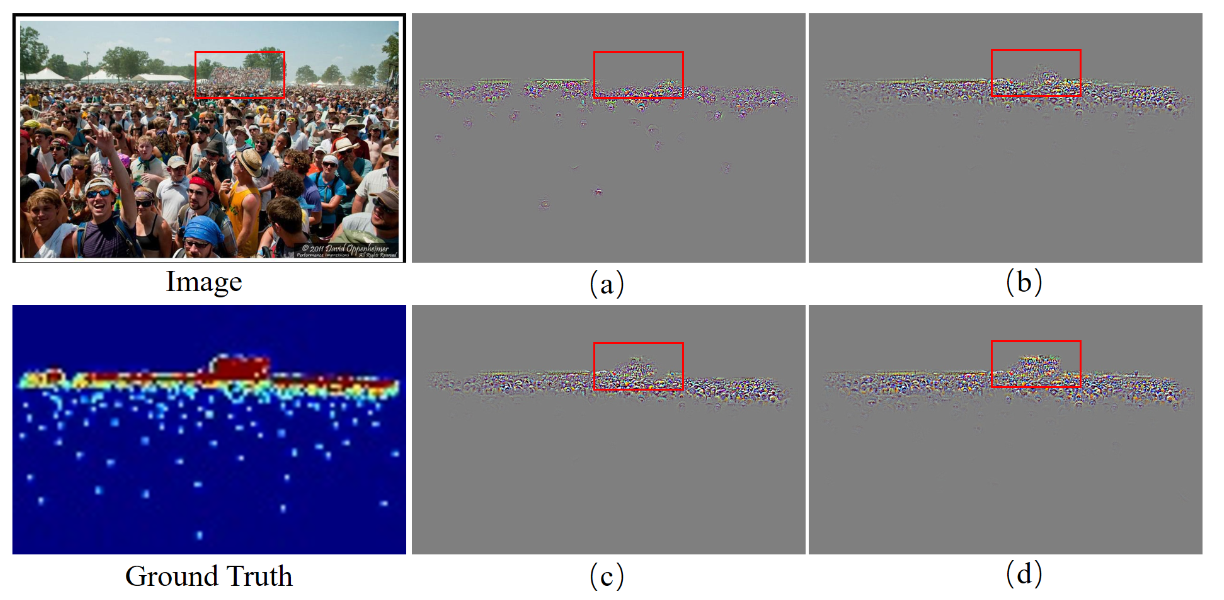}
	\caption{Feature visualization results of a sample. (a), (b), (c), and (d) represent the visualization results of the first convolution layer of the sixth convolution block by adopting the Baseline, Base$+$SH, Base$+$SH$+$SK, and Base$+$SH$+$SK$+$PLoss models, respectively.}
	\label{fig5}
\end{figure}

\textbf{ShortAgg.} As shown in Table~\ref{tableAblation}, it is noted that the network with ShortAgg (SH) module (denoted by Base$+$SH ) improves the counting performance, which shows that ShortAgg effectively promotes the learned CNN features. Specifically, ShortAgg improves the counting accuracy by 8.7$\%$ and 6.4$\%$ in terms of MAE and MSE, respectively. Furthermore, we present a comparison of the feature visualization of the same convolutional layer between the baseline and the Base$+$SH network in (a) and (b) of Figure~\ref{fig5}. We mark the high-density crowd region far away from the camera with a red rectangle. It is noted that in the red rectangle region, there is almost no feature response in the baseline model. In contrast, the Base$+$SH model has a stronger feature response.

\begin{table}[tbh]
	\centering
	\caption{Ablation study of the proposed ShortAgg (SH), SkipAgg (SK), and PLoss on ShanghaiTech Part A dataset~\cite{zhang2016single}.}
	\label{tableAblation}
	\scalebox{1.0}{
		\begin{tabular}{|c|c|c|}
			\hline
			& \multicolumn{2}{c|}{Part A}\\
			\hline
			
			Method & MAE & MSE\\
			\hline
			\hline
			Baseline & 65.22 & 102.18\\
		
			Base$+$SH & 59.54 & 95.61\\
		
			Base$+$SH$+$SK & 56.18 & 92.19\\
			
			Base$+$SH$+$SK$+$PLoss & \textbf{54.67} & \textbf{87.66}\\
			\hline
		\end{tabular}\medskip{}}
	\vspace{-0.3cm}
\end{table}

\begin{figure}[tbh]
	\centering
	\includegraphics[scale=0.35]{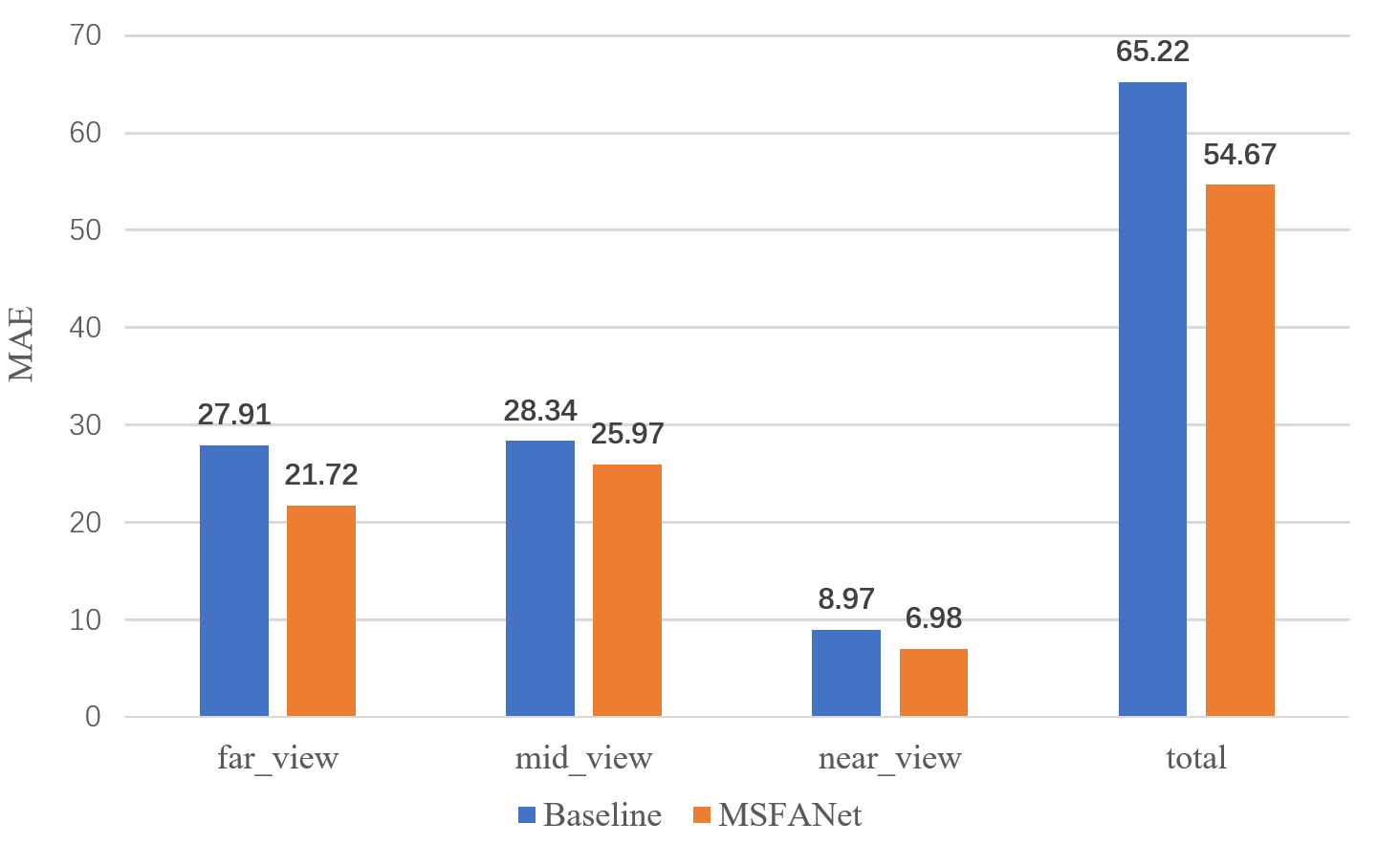}
	\caption{Statistics the counting errors(MAE) of baseline and MSFANet in three different distance regions of the ShanghaiTech part A test set.} 
	\label{fig6}
\end{figure}

\textbf{SkipAgg.} It is seen in Table~\ref{tableAblation} that the SkipAgg (SK) module further improves the Base$+$SH network. Compared with the Base$+$SH network, the obtained Base$+$SH$+$SK network decreases the MAE from 59.54 to 56.18 and decreases the MSE from 95.61 to 92.19, respectively. This means that SkipAgg effectively merges the local low-level transformer features with the high-level CNN features and makes the output features better represent the small people's heads far away from the camera. This is also confirmed by the results of the feature visualization presented in b and c of Figure~\ref{fig5}. It is seen that the SkipAgg module makes the network has much stronger feature responses in the red rectangle region.

\textbf{PLoss.} In the above ablation study, the Baseline, Base$+$SH, and Base$+$SH$+$SK models only adopt the Euclidean distance loss to train the model. 
Now we use the combination of PLoss and standard Euclidean distance loss $L_T$ to supervise the MSFANet training. It can be seen from Table~\ref{tableAblation} that the combined loss decreases the MAE from 56.18 to 54.67 on Shanghai A dataset. In our experiments, for simplicity, we adopt the non-overlapping strategy when computing the PLoss. That is, the pooling window size equals the sliding stride and we set window size to $4\times 4$ in all the experiments.

All in all, it can be seen from the red rectangle region in Figure~\ref{fig5} that the feature response of the Base$+$SH$+$SK$+$PLoss model is closer to the distribution of the ground truth, which demonstrates that the proposed method has better feature representation ability for high-density regions far away from the camera. 

Moreover, we present the statistical analysis and visualization analysis to further testify the effectiveness of the proposed MSFANet method. We divide an image scene into three regions according to the perspective, which is denoted by far\_view, mid\_view, and near\_view, respectively. Typically, the people densities in the far\_view and mid\_view regions (regions far away from the camera) are larger than those in the near\_view region. 

\begin{figure}[tbh]
	\centering
	\includegraphics[scale=0.26]{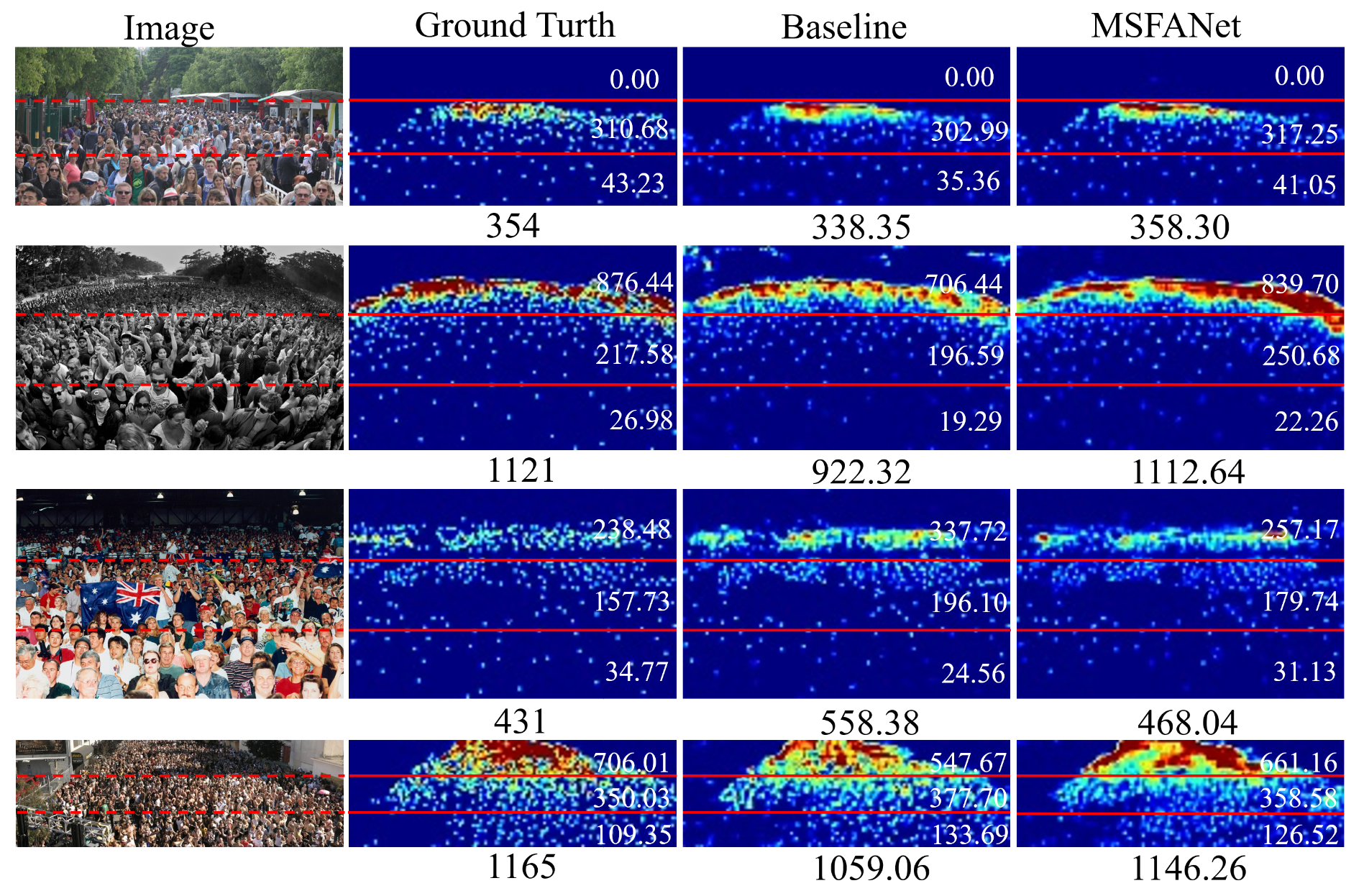}
	\caption{Visualization analysis on the ShanghaiTech Part A dataset. Compared with the baseline network, MSFANet effectively reduces the density estimation errors in different scenes, especially the high-density regions far away from the camera.}
	\label{fig7}
\end{figure}
As presented in Figure~\ref{fig6}, compared with the baseline network, the MSFANet method indeed reduces the counting errors of different distance regions, especially in the far\_view and mid\_view regions. Specifically our MSFANet reduces the MAEs of the far\_view and mid\_view regions by 20\% and 8.4\% respectively.
Examples of the visualization analysis between the baseline network and our MSFANet on ShanghaiTech Part A are presented in Figure~\ref{fig7}. It can be seen that our MSFANet has much more accurate density estimations in all three regions.

\subsection{Results and Analysis}
We evaluate the proposed MSFANet method against eighteen state-of-the-art  methods~\cite{zhang2019relational,LI2020106485,GUO2021106691,jiang2019crowd,miao2020shallow,liu2019context,yang2020reverse,yan2019perspective,luo2020hybrid,wan2021generalized,xiong2019open,liu2020adaptive,hu2020count,abousamra2021localization,ma2021learning,jiang2020attention,liu2020weighing,bai2020adaptive} on ShanghaiTech Part A ~\cite{zhang2016single}, UCF\_CC\_50~\cite{idrees2013multi} ,UCF-QNRF~\cite{idrees2018composition}, and  WorldExpo'10~\cite{zhang2015cross} datasets.We use the MAE metric and MSE metric to evaluate these method. In addition, we rank these methods according to MAE.The detailed experimental results are presented in Table~\ref{tableResult}. 

\textbf{ShanghaiTech Part A Dataset.}
For simplicity, SHTech PartA represents the ShanghaiTech Part A Dataset  in Table~\ref{tableResult}.On this dataset, our method achieves the best result with an MAE of 54.67, which is 1.3\% lower than ADSCNet~\cite{bai2020adaptive}.

\textbf{UCF\_CC\_50 Dataset.}
Following the standard protocol adopted in ~\cite{idrees2013multi}, we perform 5-fold cross-validation to evaluate the proposed MSFANet on UCF\_CC\_50 Dataset. As shown in Table~\ref{tableResult}, it is observed that our MSFANet outperforms all other methods with the lowest MAE of 159.1 ,which is 9.0\% lower than that of the second-best ASNet~\cite{jiang2020attention}. 

\textbf {UCF-QNRF Dataset.}
On the UCF-QNRF Dataset, since most of the original images are very large, we resize the larger side of the input image to 2048 pixels and keep the aspect ratio of the image unchanged. Our method achieves an MAE of 86.24 that ranks 4th among the state-of-the-art methods.

\begin{table*}[t!]
	\begin{center}
		\resizebox{\textwidth}{!}{
			\setlength{\tabcolsep}{3pt}
			\begin{tabular}{|l|ccc|ccc|ccc|ccccccc|c|}
				\hline
				&\multicolumn{3}{|c|}{SHTech Part A}&\multicolumn{3}{|c|}{UCF\_CC\_50}
				&\multicolumn{3}{|c|}{UCF-QNRF}
				&\multicolumn{7}{|c|}{WorldExpo10}&\\
				\hline
				Method &MAE&MSE&R.
				&MAE&MSE&R. 
				&MAE&MSE&R.
				&S1&S2&S3&S4&S5&Avg.
				&R.&avg. R.\\ 				
				\hline
				\hline

				TEDnet~\cite{jiang2019crowd} &64.2 &109.1 &18
				&249.4  &354.5  &14
				&113    &188    &15
				&2.3 &10.1   &11.3   &13.8   &2.6  &8.0  &8 
				&13.75 \\
				
				DSA-Net~\cite{LI2020106485} &67.4 &104.6 &19
				&-  &-  &-
				&-    &-    &-
				&2.4 &11.2   &12.7   &9.5   &2.4  &7.6  &7
				&13 \\

				SDANet~\cite{miao2020shallow} &63.6  &101.8  &17 
				&227.6  &316.4 &11
				&-&-&- 
				&2.0   &14.3   &12.5  &9.5  &\textbf{2.5}  &8.16  &10 
				&12.67 \\
				
				RANet~\cite{zhang2019relational}&59.4 &102.0 &10 
				&239.8 &319.4 &12
				&111 &190 &14
				&-&-&-&-&-&-&-
				&12 \\

				CAN~\cite{liu2019context}&62.3  &100.0  &16
				&212.2  &243.7  & 10
				&107   &183    &13
				&2.9   &12.0   &10.0  &7.9  &4.3  &7.4  &6
				&10.5 \\
				
				RPNet~\cite{yang2020reverse} &61.2  &  91.9  &12 
				&-&-&- 
				&-&-&- 
				&2.4   &10.2  &9.7  &11.5  &3.8  &8.2  &11
				&11.5 \\

				PGCNet~\cite{yan2019perspective} &57.0 &\textbf{86.0} &6
				&244.6 &361.2 &13
				&-&-&-
				&2.5 &12.7 &\textbf{8.4} &13.7 &3.2 &8.1 &9
				&9.33 \\
				
				HyGnn~\cite{luo2020hybrid} &60.2 &94.5  &11
				&184.4  &270.1 &6
				&100.8  &185.3 &10
				&-&-&-&-&-&-&-
				&9 \\
				
				TopoCount~\cite{abousamra2021localization}&61.2  &104.6 &13
				&184.1  & 258.3  &5
				&89.0   & 159.0  &7
				&-&-&-&-&-&-&- 
				&8.33 \\
				
				GLoss~\cite{wan2021generalized}&61.3 &95.4 &14
				&-&-&- 
				&84.3   &147.5  &3
				&-&-&-&-&-&-&- 
				&8.5 \\
				
				AMRNet~\cite{liu2020adaptive} &61.59  &98.36  &15 
				&184.0   &265.8  & 4
				&86.6   &152.2  &5
				&-&-&-&-&-&-&- 
				&8\\
				
				S-DCNet~\cite{xiong2019open}&58.3  &95.0  &9
				&204.2  &301.3  &8
				&104.4   &176.1  &12 
				&\textbf{1.57}  &9.51  &9.46  &10.35  &2.49  &6.67  &3 
				&8 \\

				AMSNet~\cite{hu2020count} &56.7  &93.4  &5
				&208.4   &297.3  &9
				&101.8   &163.2   &11
				&1.6     &\textbf{8.8}    &10.8  &10.4 &\textbf{2.5} &6.8  &5 
				&7.5\\
				
				CHANet~\cite{GUO2021106691} &55.8 &95.6 &3
				&-  &-  &-
				&98    &177    &9
				&1.4 &11.6   &8.7   &8.9   &2.8  &6.7  &4
				&5.33 \\

				UOT~\cite{ma2021learning} &58.1 &95.9 &8
				&-&-&- 
				&83.3   &142.3  &2
				&-&-&-&-&-&-&- 
				&5 \\
				
				ASNet~\cite{jiang2020attention} &57.78  &90.13  &7
				&174.84  &251.63  & 2
				&91.59   &159.71  &8
				&2.22   &10.11   &8.89  &\textbf{7.14}  &4.84  &6.64   &2 
				&4.75 \\
				
				LibraNet~\cite{liu2020weighing} &55.9  &97.1  &4 
				&181.2   &262.2   &3
				&88.1   &143.7    &6
				&-&-&-&-&-&-&- 
				&4.33\\
				
				ADSCNet~\cite{bai2020adaptive} &55.4  &97.7  &2 
				&198.4   &267.3 &7
				&\textbf{71.3}   &\textbf{132.5}  &\textbf{1}
				&-&-&-&-&-&-&- 
				&3.33\\
				
				MSFANet   & \textbf{54.67} & 89.89 &\textbf{1}  
				& \textbf{159.1} & \textbf{230.6}&\textbf{1} 
				& 86.24 & 148.20 &4
				&1.59 & 11.25 & 8.92 & 8.50 & 2.6 & \textbf{6.57} &\textbf{1} 
				&\textbf{1.75}\\
				\hline
			\end{tabular}
		}\vspace{-0.3cm}
	\end{center}
	\caption{Comparison of the proposed MSFANet with other state-of-the-art methods on four challenging datasets. The average ranking performance (denoted by Avg. R.) is obtained by using the sum of all rankings that one method gains to divide the number of datasets it utilizes. Avg. R. can demonstrate the counting performance of the method across different crowd scenes.}
	\vspace{-0.3cm}
	\label{tableResult}
\end{table*}

This dataset provides Region of Interest (ROI) along with perspective maps for both training and test scenes.

\textbf{WorldExpo'10 Dataset.}
The WorldExpo'10 dataset \cite{zhang2015cross} provides Region of Interest (ROI) along with perspective maps for both training and test scenes. To be consistent with previous
work~\cite{zhang2015cross},we prune the last convolutional layer so that the features out of ROI regions are set to zero. In addition,only the MAE metric is used to evaluate the results.
First, the MAE of each test scenes is calculated, and then all the results are averaged to evaluate the performance of MSFANet across different test scenes.It is observed that the proposed MSFANet outperforms other approaches with the lowest average MAE of 6.57 while achieving competitive performance in each test scene.

By following the ASNet~\cite{jiang2020attention}, we also adopt the average ranking metric to comprehensively evaluate the proposed methods.Avg. R. represents the average ranking  in Table~\ref{tableResult}, which is obtained by calculating the sum of the ranking of this method on each dataset divided by the number of datasets.It is noted that in Table~\ref{tableResult} that our method ranks first in the average ranking.

The average ranking is obtained by calculating the sum of the ranking of this method on each data set divided by the number of data sets, in which the ranking of unpublished data sets does not participate in the calculation.

\section{Conclusion}

In this paper, we have proposed a novel multi-scale feature aggregation network (MSFANet) that merges features of different receptive fields to reduce the density estimation errors, especially in high-density regions. To this end, we introduce two feature aggregation modules named short aggregation (ShortAgg) and skip aggregation (SkipAgg) to jointly enhance the feature representation ability of the counting network.
ShortAgg gradually merges features of the adjacent scales and SkipAgg directly propagates the low-level transformer features to the high-level CNN features. The combination of ShortAgg and SkipAgg makes the output features of the MSFANet robust to different crowd distributions, especially the high-density scenes. 
Furthermore, we introduce the global-and-local counting loss by combining the standard Euclidean distance loss and the proposed pooling Loss (PLoss). PLoss utilizes a locality-aware loss kernel to generate the counting loss in the pooling manner, so as to alleviate the impact of the non-uniform crowd data on the training procedure.
Extensive experiments on four challenging datasets demonstrate that the proposed method achieves promising results via the two easy-to-implement feature aggregation modules.


\bibliographystyle{elsarticle-num}
\bibliography{ref}


%
%
%

\end{document}